\documentclass[review]{elsarticle}

\usepackage{lineno,hyperref}
\usepackage{epsfig}
\usepackage{graphicx}
\usepackage{multirow}
\usepackage{booktabs}
\usepackage{amsmath,amssymb}
\usepackage{color}
\newcommand{\etal}{\textit{et al.}}

\journal{Journal of Pattern Recognition Templates}

\begin{document}

\begin{frontmatter}

\title{Towards Disentangled Representations for Human Retargeting by Multi-view Learning}

\author[my1address]{Chao Yang}

\author[my2address]{Xiaofeng Liu}

\author[my3address]{Qingming Tang}

\author[my1address]{C.-C. Jay Kuo}

\address[my1address]{Department of Computer Science, University of Southern California, Los Angeles, CA, 90089 USA}
\address[my2address]{Harvard University, Cambridge, MA, 02138 USA}

\address[my3address]{Toyota Technological Institute at Chicago, IL, 60637 USA}

\begin{abstract}

We study the problem of learning disentangled representations for data across multiple domains and its applications in human retargeting. Our goal is to map an input image to an identity-invariant latent representation that captures intrinsic factors such as expressions and poses. To this end, we present a novel multi-view learning approach that leverages various data sources such as images, keypoints, and poses. Our model consists of multiple id-conditioned VAEs for different views of the data. During training, we encourage the latent embeddings to be consistent across these views. Our observation is that auxiliary data like keypoints and poses contain critical, id-agnostic semantic information, and it is easier to train a disentangling CVAE on these simpler views to separate such semantics from other id-specific attributes. We show that training multi-view CVAEs and encourage latent-consistency guides the image encoding to preserve the semantics of expressions and poses, leading to improved disentangled representations and better human retargeting results.  

~\\~

\end{abstract}

\begin{keyword}
\texttt{Disentanglement}\sep \texttt{Retargeting} \sep \texttt{Multi-view}

\end{keyword}

\end{frontmatter}


~\\~

\section{Introduction}

\begin{figure}[!h]
\centering
\small
\setlength{\tabcolsep}{1pt}
\begin{tabular}{c@{\hskip .13in}cccc}
\includegraphics[width=.13\textwidth]{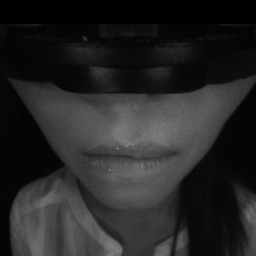}&
  \includegraphics[width=.13\textwidth]{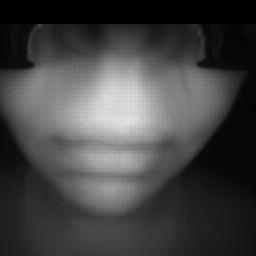}&
  \includegraphics[width=.13\textwidth]{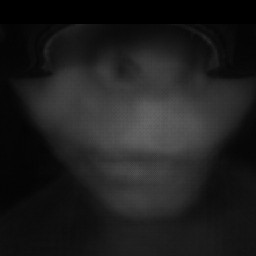}&
  \includegraphics[width=.13\textwidth]{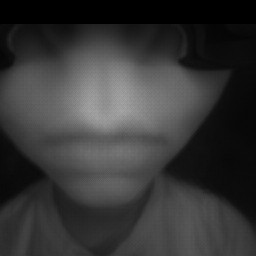}&
  \includegraphics[width=.13\textwidth]{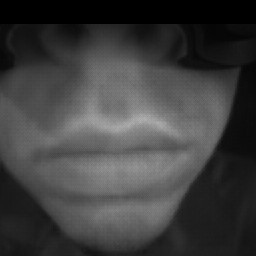}\\
\includegraphics[width=.13\textwidth]{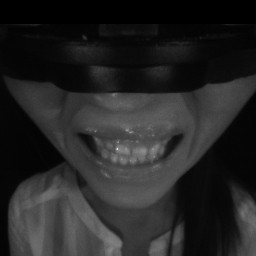}&
  \includegraphics[width=.13\textwidth]{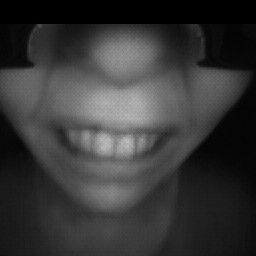}&
  \includegraphics[width=.13\textwidth]{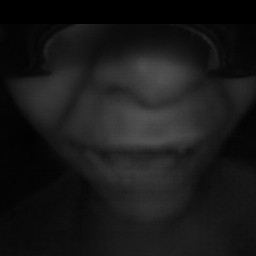}&
  \includegraphics[width=.13\textwidth]{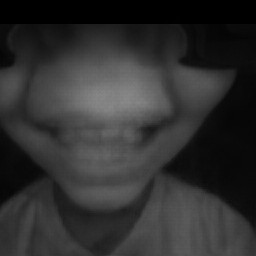}&
  \includegraphics[width=.13\textwidth]{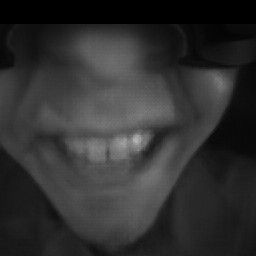}\\
  \includegraphics[width=.13\textwidth]{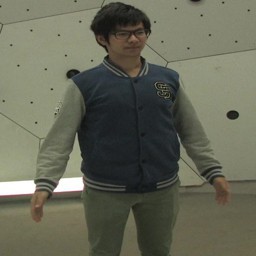}&
  \includegraphics[width=.13\textwidth]{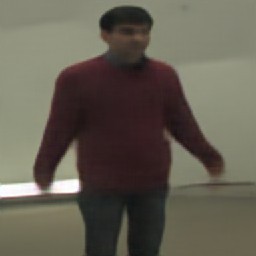}&
  \includegraphics[width=.13\textwidth]{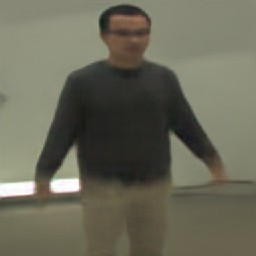}&
  \includegraphics[width=.13\textwidth]{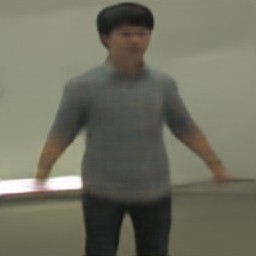}&
  \includegraphics[width=.13\textwidth]{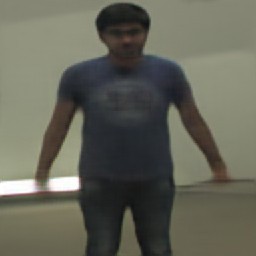}\\
  \includegraphics[width=.13\textwidth]{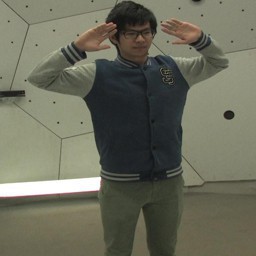}&
  \includegraphics[width=.13\textwidth]{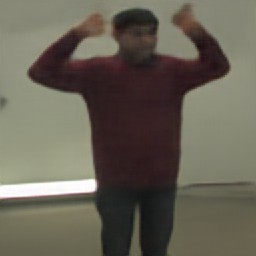}&
  \includegraphics[width=.13\textwidth]{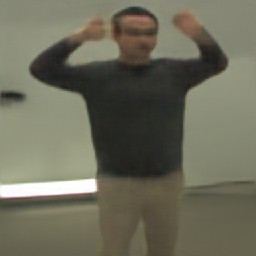}&
  \includegraphics[width=.13\textwidth]{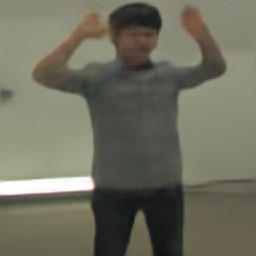}&
  \includegraphics[width=.13\textwidth]{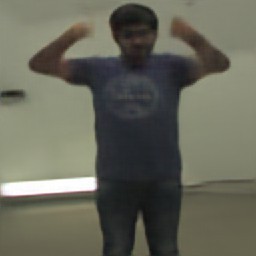}\\
\end{tabular}
\caption{Visual results of face (with head-mounted displays) and body retargeting. The leftmost column is the input, and the right four columns are the output after changing the identity labels. After learning the disentangled representations of the expressions and poses from unannotated data, we can switch the identities and keep the expressions or poses the same.}.
\label{fig:teaser}
\end{figure}

We consider the problem of learning disentangled representations for sensory data across multiple domains. Ideally, such representations are interpretable and can separate factors of variations. Given such factors could be domain-specific (such as identities, categories, etc.) and domain-invariant (such as shared expressions, poses, illuminations, etc.), a simpler (and more practical) question to ask is how to separate the domain-invariant factors from the domain-specific factors. This could potentially enable cross-domain image translations or in-domain attribute interpolation/manipulations. The problem then boils down to learning an underlying representation that is invariant to domain changes, yet encompasses the entirety of other factors that describe the data. Unfortunately, without explicit correspondences or prior knowledge, such representations are inherently ambiguous~\cite{szabo2017challenges}.

In this work, we are specifically interested in learning identity-invariant representations from human captures. The data could come from different sources such as cameras~\cite{lombardi2018deep} or RGBD sensors~\cite{Joo_2017_TPAMI}. Learning identity-invariant representations is one of the critical steps towards understanding the underlying factors describing human appearances and movements, and could foster applications on multiple fronts. In more detail, our motivations to tackle this problem are i) such datasets are common and relatively easy to acquire without requiring additional annotations, and can serve as ideal test-beds for generative models. ii) it enables human face/body retargeting and synthesis which have interesting real-world applications~\cite{thies2016face2face} and iii) The learned identity-invariant representations can be used to train other down-stream tasks such as expression/pose detection or 3D avatar driving~\cite{lombardi2018deep}. 

Recently, deep learning techniques have led to highly successful image translations and feature disentanglement. Many existing works address cross-domain image translations without explicitly modeling the underlying factors that explain the images~\cite{zhu2017unpaired,wang2017high}. The limitation is that it is impossible to directly manipulate the attributes or use the learned representations as a proxy for other tasks. Other techniques that jointly learn representation disentanglement and image translation require paired data during training~\cite{gonzalez2018image} to marginalize and factor out independent factors. Several works have been proposed to eliminate the need for paired training data by using adversarial training or cycle-consistency assumption. However, there is little guarantee that we separate content and style as desired as the networks could lose essential intra-domain information~\cite{lee2018diverse}. One of the most similar work to ours is~\cite{liu2018unified}, which augments a CVAE with adversarial training to achieve multi-domain image translation and learn domain-invariant representations. However, our experiments show that intra-domain semantic information such as poses or expressions can be easily lost if we apply adversarial loss on the latent space.

We observe that a successful representations should meet two requirements. First, it needs to capture intra-domain variances between data within the same domain, and second, it needs to ignore inter-domain variances of data across different domains. In the case of face retargeting, this means that different people making the same expressions should have identical (or very similar) representations, while different expressions should have different representations. 
A straignth-forward approach is to learn a conditional variational auto-encoder (CVAE)~\cite{sohn2015learning} to extract representations that conditioned on domain labels. 
However, there is no guarantee that the conditional representations learned by CVAE would tend to leave out domain information.
Actually, 
the Kullback-Leibler divergence between the conditional posterior and unconditional prior in CVAE trades off domain-invariant consistency against fidelity of reconstruction ~\cite{higgins2017beta}.
We observe that this trade-off becomes a more severe issue when augmenting the formulation with adverserial regularization~\cite{liu2018unified}.

To address this challenge, we notice that for many of the real-world problems, we often have access to multi-view information that can be used. For example, it is easy to detect facial keypoints or body poses from images. The advantage of such auxiliary information is that, by construction, they already encoder semantic information about the visual phenomena as they are human defined and annotated. We show that it is easier to learn identity-invariant representations from the simplified data. However, the identity-invariant representations learned from keypoints or poses alone do not contain all the information needed to reconstruct the images -- we may lose information like the facial details or global illuminations. Therefore, it is critical to extract information from both the image data and auxiliary data. Our core idea is to leverage the auxiliary data to simplify the task of learning domain-invariant representations from the images. We propose two variants of models that can achieve this goal, showing that using multi-view information such as keypoints or poses are useful, and can lead to better disentangled representations and image reconstructions.

We demonstrate the effectiveness of our approach on several datasets. The first one is the large-scale head-mounted captures (HMC) data that we collected. The data consists of video sequences of more than 100 identities wearing virtual reality (VR) headsets, each making a variety of facial expressions. The goal is to learn an identity-invariant representation that encodes the facial expressions. This can be applied to drive 3D avatars in VR in real time when wearing the headset. We also test our algorithm on the CMU Panoptic dataset~\cite{Joo_2017_TPAMI} that consists of 30 people making different poses. Similarly, the goal is to learn an identity-invariant representation that contains body pose information. We evaluate our approach by plotting t-SNE~\cite{maaten2008visualizing} of our latent code and also measuring the reconstruction quality and the semantic correctness. We also compare our method against several baselines and existing techniques to show its advantage.

We summarize our contributions as following:
\begin{enumerate}
	\item We propose two simple CVAE-based frameworks that use multi-view information to learn domain-invariant representations. 
	\begin{enumerate}
	    \item Our approach can learn domain invariant representation significantly better than alternatives we have tried.
	    \item Our approach can also be understood as distilling the knowledge hidden in ``generating key-points" to a deep neural network, thus achieving the goal of learning id-invariant representation.
	    \item Our approach (and also our implementation of CVAE) uses domain labels as privileged information. By this way, for the unseen images without domain labels in the future, our inference is still capable to learn representations.
	\end{enumerate}
	\item We enable large-scale image translations of multiple domains that achieve state-of-the-art results. 
	\item We show several novel applications in human retargeting and VR.
\end{enumerate}

\section{Related Work}

\noindent\textbf{Disentangled Representation} The problem of learning disentangled representation has long been studied with vast literature. Before deep learning,~\cite{tenenbaum1997separating} uses bilinear models to separate content and style and~\cite{huang2007unsupervised} uses convolution filters to learn a hierarchy of sparse feature that is invariant to small shifts and distortions. A lot of recent developments on representation disentanglement are based on models like generative adversarial network (GAN)~\cite{goodfellow2014generative} and variational auto-encoders (VAE)~\cite{kingma2013auto}. Representative works include pixel-level domain separation~\cite{bousmalis2016domain} and adaptation~\cite{bousmalis2017unsupervised}, video disentanglement~\cite{tulyakov2017mocogan} and 3D graphics rendering~\cite{kulkarni2015deep}. However, many of the existing works require supervision or semi-supervision~\cite{kingma2014semi}. InfoGAN~\cite{chen2016infogan} proposes to learn disentangled representations in a completely unsupervised manner, although is limited to learn disentangled representation in a singled domain and cannot be easily extended to describe cross-domain data. In beta-VAE~\cite{higgins2017beta}, it modifies the VAE framework, and introduce an adjustable hyper-parameter beta that balances reconstruction accuracy and compactness of the latent code. Nevertheless, it is still hard to achieve satisfying disentangling performance.~\cite{szabo2017challenges} improve the performance of unsupervised disentanglement by applying adversarial training in the output space. We found while it usually improves the quality of generated images but does not learn disentangled representations better. One of the most similar works to ours is~\cite{liu2018unified}, which is based on a conditional VAE (CVAE) trained across multiple domains, with adversarial loss applied to the latent space. We notice that while it is effective in making the latent code more domain-invariant, it makes the reconstruction less accurate as the number of hidden factors increases.

\noindent\textbf{Image Translation} One of the main applications of our disentangled representation is for cross-domain image translations. Deep learning-based image translation has motivated many recent research. Isola \etal~\cite{isola2017image} proposes Pix2Pix, which uses conditional GANs and paired training data for image translation between two domains. As follow-up works, Wang \etal~\cite{wang2018pix2pixHD} extended the framework to generate high-resolution images conditioning on the source domain; BicycleGAN~\cite{zhu2017toward} adds noise as input and applies it to multimodal image translation. Without paired training data as supervision, additional constraints are used to regularize the translation. For example, CycleGAN~\cite{zhu2017unpaired} proposes to use cycle-consistency to ensure 1-1 mapping, which generates impressive results on many datasets. Similar ideas have been applied by Kim \etal~\cite{kim2017learning}, Yi \etal~\cite{yi2017dualgan} and Liu \etal~\cite{liu2016coupled}. StarGAN~\cite{choi2017stargan} recently proposes to extend this framework to multi-domain image translation and shows it could switch the diverse attributes of human faces without annotating the correspondences. However, these methods do not explicitly learn disentangled latent representations and the applications are limited. Other works tackle the problem of image translation and disentangled representation jointly, mostly relying on adversarial training~\cite{rifai2012disentangling} or cycle-consistency constraints~\cite{lee2018diverse}. However, they are either limited to dual domains or suffer from the aforementioned reconstruction-disentanglement trade-off. 

\begin{figure}[t]
	\centering
	\includegraphics[width=.9\linewidth]{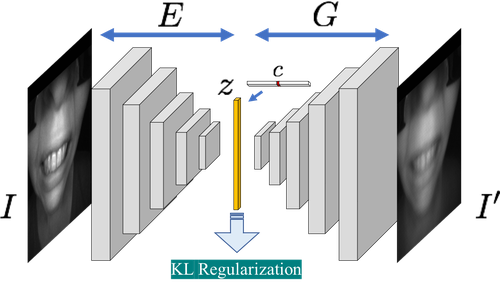}
	\caption{The baseline CVAE model. The decoder is conditioned on identity $c$, which is encoded as 1-hot vector.} 
\end{figure}
\label{fig:fwb}

\noindent\textbf{Human Retargeting} Face retargeting and reenactment attracts a lot research interest given its wide application~\cite{dale2011video}. Similar to our paper, Olszewski \etal~\cite{olszewski2016high} studies the facial and speech animation in the setting of VR head-mounted displays (HMDs). Face2Face~\cite{thies2016face2face} achieves real-time face capture and reenactment through online tracking and expression transfer. The quality of the facial textures is further improved using conditional GANs in~\cite{olszewski2017realistic}. Similarly,~\cite{korshunova2017fast} achieves face swapping using multi-scale neural networks. For body image synthesis and transfer, ~\cite{ma2017pose} applies a two-step process, which generates the global structure first and then synthesizes fine details. ~\cite{esser2018variational} devises a framework based on variational u-net that separates appearances and poses. However, it requires a neutral representation of the target identity as input and is unable to achieve retargeting directly.

\section{Our Approach}

\begin{figure*}[t]
    \centering
    \includegraphics[width=.49\linewidth]{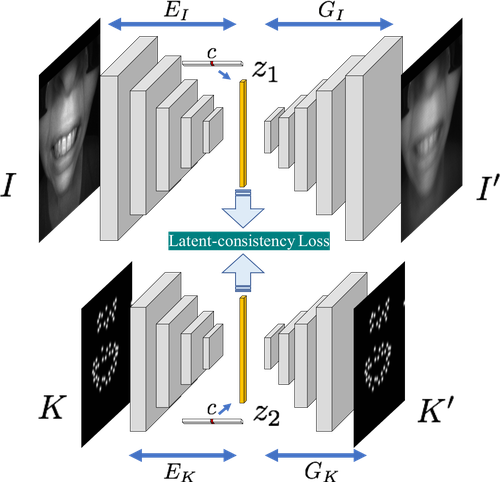}
    \includegraphics[width=.49\linewidth]{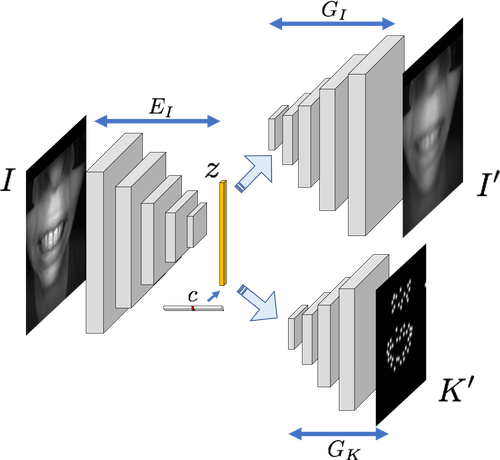}
    \caption{Two variants of our models. Left: jointly train two CVAEs for images and keypoints and enforce latent-consistency constraint. Right: train a single CVAE but generate the keypoints alongside the image.} 
\end{figure*}
\label{fig:framework}

\subsection{Learning Domain-Invariant Representations}

We first define the problem of learning domain-invariant representations. 
Given a collection of images $\mathcal{X}=\{\mathcal{X}_c\}^N_{c=1}$ across $N$ domains ($N$ identities here), 
the task of learning domain-invariant representations is to learn a low-dimensional embedding $z$ for each image $x \in 
\mathcal{X}$, %
where $z$ should capture the semantics shared across identities such as poses and expressions, 
while disregarding identity-specific attributes.
When we know the semantics explicitly, 
or we have the correspondences of semantics between different domains as supervision (e.g. two people with same pose expression), 
we can directly infer $z$ to be consistent with the semantics. 
However, acquiring such supervision is usually expensive. 
Instead, we are interested in the setting where no explicit supervision is available, 
except we know the domain label (identity) of each image. 

\subsection{Baseline: Conditional Variational Autoencoder}
Conditional Variational Autoencoder (CVAE) is a type of deep generative models, 
which aims to form probabilistic representations that explains the data given the domain labels. 
Specific to our implementation, for a sample $x$, 
CVAE learns low-dimensional embedding $z$ that maximizes the log-likelihood $\log p(x|c)=\log \int_z(p(x|z,c)p(z|c))$,
which is typically intractable.
Thus we maximize the Evidence Lower Bound (ELBO)~\eqref{eqn:elbo} instead: 
\small
\begin{align}
E_{q_{\phi}(z|x)}[\log p_{\theta}(x|z,c)] - D_{\mathcal{KL}}(q_{\phi} (z|x)|| p_{\theta}(z|c))
\label{eqn:elbo}
\end{align}
\normalsize

In \eqref{eqn:elbo}, $\theta$ are the model parameters of decoder $G$ (generation network), 
and $\phi$ are the model parameters of encoder $E$ (inference network), 
and $c$ is a specific identity. 
$p_{\theta}(z|c)$ is the prior distribution (we use $p_{\theta}(z|c) = \mathcal{N}(0,1)$) and $q_{\phi}(z|x)$ is the variational posterior aims at approximating $p_{\theta}(z|x,c)$. 
We maximize ~\eqref{eqn:elbo} by minimizing the following loss function ~\eqref{eqn:new-elbo} :
\small
\begin{align}
\mathcal{L}(\theta,\phi) = E_{q_{\phi}(z|x)} \big\{ \left\| G(z,c)-x\right\|_1 \big\} + D_{\mathcal{KL}} ( \mathcal{N}(\mu(x), \sigma^2(x) I) || \mathcal{N}(0,I))
\label{eqn:new-elbo}
\end{align}
\normalsize
where we denote
$q_{\phi}(z|x)= \mathcal{N}(\mu(x), \sigma^2(x) I)$ is a diagonal Gaussian
and
$p_{\theta}(x|z,c) $ proportional to $e^{- \left\| G(z,c)-x\right\|_1}$.
After training a plain CVAE as shown in \eqref{eqn:new-elbo}, 
the inference network can create representation $z$ of input image $x$ without using identity $c$,
as we only concatenate identity $c$ with samples from latent variable $z$ towards reconstruction, 
so identity $c$ is actually used as privileged information during learning.

Ideally, we wish to learn latent code $z \sim 
\mathcal{N}(\mu(x), \sigma^2(x))$ only pertinent to domain-invariant factor $d_x$. This is usually addressed by several techniques: i) increase the weight of $D_{\mathcal{KL}}$ regularization~\cite{higgins2017beta}; ii) reduce the dimensionality of $z$~\cite{lee2018diverse} or iii) apply domain-adversarial loss on $z$~\cite{szabo2017challenges}. All of these can lead to a more compact representation of $z$ and tends to leave out information only correlated with class label $c$. However, we experimentally noticed that the major drawback is that they also limit the ability of $z$ to fully encode $d_x$ and model the intra-domain variances, hence the reconstruction quality suffers. Also, it may not be convenient as it requires time-consuming trial-and-error procedures to find the appropriate hyper-parameters. We show these challenges could be alleviated by leveraging multi-view information from the data.

\subsection{Disentangling with Multi-view Information}

\begin{figure}[!h]
\centering
\small
\setlength{\tabcolsep}{1pt}
\begin{tabular}{c@{\hskip .13in}cccc}
\includegraphics[width=.13\textwidth]{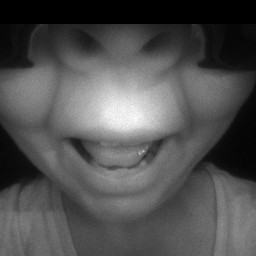}&
  \includegraphics[width=.13\textwidth]{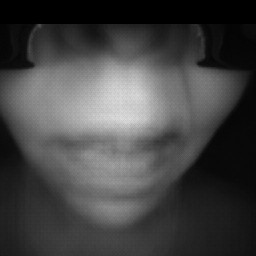}&
  \includegraphics[width=.13\textwidth]{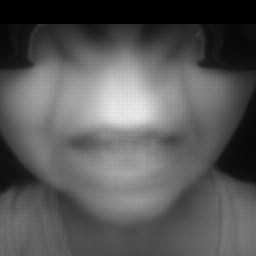}&
  \includegraphics[width=.13\textwidth]{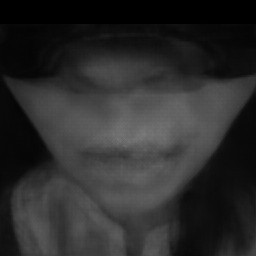}&
  \includegraphics[width=.13\textwidth]{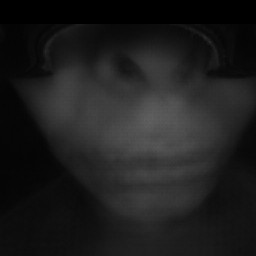}\\
\includegraphics[width=.13\textwidth]{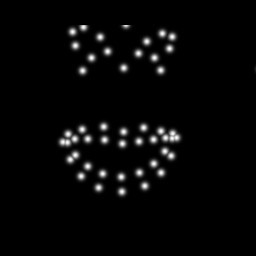}&
  \includegraphics[width=.13\textwidth]{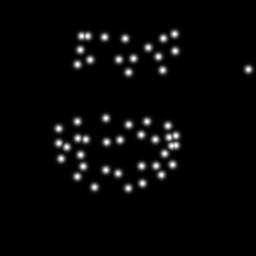}&
  \includegraphics[width=.13\textwidth]{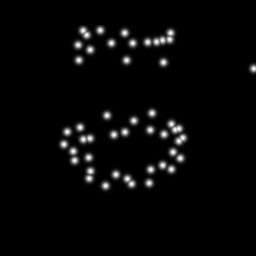}&
  \includegraphics[width=.13\textwidth]{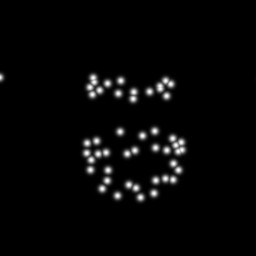}&
  \includegraphics[width=.13\textwidth]{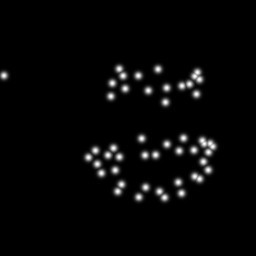}\\
  \includegraphics[width=.13\textwidth]{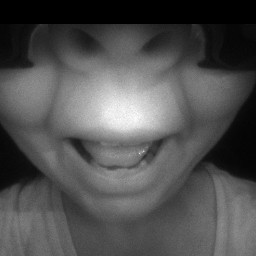}&
  \includegraphics[width=.13\textwidth]{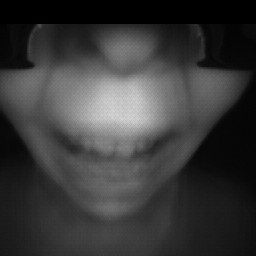}&
  \includegraphics[width=.13\textwidth]{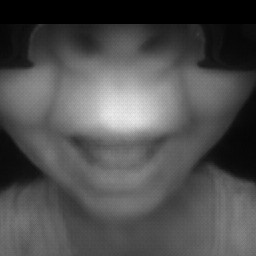}&
  \includegraphics[width=.13\textwidth]{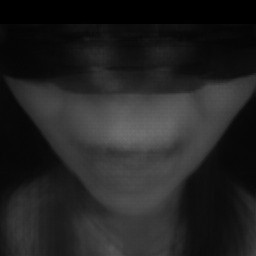}&
  \includegraphics[width=.13\textwidth]{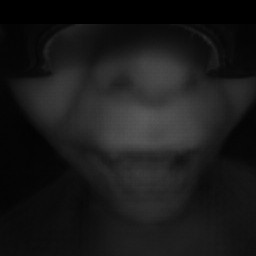}\\
\end{tabular}
\caption{Top: results of image-based CVAE baseline. Middle: results of keypoints-based CVAE. Bottom: results of jointly train image and keypoints with latent-consistency constraint. We can see for this example the image-based CVAE fails to encode the facial expression into the latent representation, while the keypoints-based CVAE successfully models and transfers the expressions. By jointly training the two CVAEs we can encourage the latent code of images to preserve the identity-invariant semantics as well.}.
\label{fig:comp}
\end{figure}

We notice a major hurdle of learning disentangled, domain-invariant representations from unannotated data is the number of factors that could explain the variations, ranging from expressions, poses, shapes, global illuminations to headset locations. Faithfully modeling all these factors while separating them from $c$ is a challenging task for training the model. On the other hand, if the variations are reduced to a small number of factors, it becomes much easier to disentangle and learn the $z$ that models $d_x$ and independent of $c$. We notice for our task of human retargeting, one of the essential intra-domain variations can be explained by facial keypoints or body poses, which are relatively easy to acquire~\cite{cao2016realtime}. 
With the keypoints, we can train a conditional VAE using the same formula~\ref{eqn:new-elbo}, with $x$ been replaced by kepoints $K$.
Fig.~\ref{fig:comp} shows an example where the keypoints-based CVAE successfully models the facial expression changes when the image-based CVAE fails. 
The latent code $z_K$ learned from keypoints could be viewed as a simplified version of the latent code $z_x$ learned from images. To model the remaining factors of variations, we could then train two CVAEs (one for keypoint and one for image) end-to-end, such that the objective function becomes:
\small
\begin{align}
\mathcal{L}(\theta,\phi)& = E_{q_{\phi}(z_K|x)} \Big\{ \left\| G_K(z_K, c)-K\right\|_1  + E_{q_{\phi}(z_x|x)} \big\{ \left\| G_I(z_x, z_K, c)-x\right\|_1 \big\} \Big\}  \nonumber \\
& + D_{\mathcal{KL}}( q_{\phi}(z_k|x) || \mathcal{N}(0,I)) + D_{\mathcal{KL}}( q_{\phi}(z_x|x)  || \mathcal{N}(0,I)) 
\label{eqn:kl2}
\end{align}
\normalsize
Unfortunately, our experiments show that this still suffers from the disentanglement-reconstruction trade-off as before. Alternatively, we propose two unified frameworks that explicitly use multi-view information as constraints and effectively leverage the simplified structure of keypoints to guide the training of image-based CVAE (Fig.~\ref{fig:framework}).

\subsubsection{Multi-view Training and Multi-view Output} 
We jointly train two CVAE models in parallel, and encourage their latent codes to be consistent by using $\left\| z_I-z_K\right\|^2$ as regularizer. The overall training objective is:

\small
\begin{align}
\mathcal{L}(\theta,\phi)& = E_{q_{\phi}(z_K|x)} \Big\{ \left\| G_K(z_K, c)-K\right\|_1  + E_{q_{\phi}(z_x|x)} \big\{ \left\| G_I(z_x, z_K, c)-x\right\|_1 \big\} \Big\}  \nonumber \\
& + \lambda_{kl} \big\{ D_{\mathcal{KL}}( q_{\phi}(z_k|x) || \mathcal{N}(0,I)) + D_{\mathcal{KL}}( q_{\phi}(z_x|x)  || \mathcal{N}(0,I)) \big\} \nonumber \\
& + \lambda_z \left\| z_x-z_K\right\|^2
\label{eqn:kl2}
\end{align}
\normalsize

As shown in Fig.~\ref{fig:framework} left, we associate the two CVAEs with the latent-consistency term, such that we explicitly encourage the latent code $z_x$ to be similar to the domain-invariant semantics within $z_K$. The hyper-parameter $\lambda_z$ affects the level of similarity between $z_x$ and $z_K$. 
If $\lambda_z$ is very large, $z_x$ would be identical to $z_K$ and is unable to model variations of factors other than keypoints semantics. On the contrary, if $\lambda_z$ is too small, it decouples the training of two branches, which leads to $z_x$ is not domain-invariant. We study the effects of $\lambda_z$ in the ablation study.

Another variant of our model is to train a single CVAE for images, but in addition to output the image reconstruction, we have a separate decoder $G_K$ that reconstructs the ground truth keypoints:
\small
\begin{align}
\mathcal{L}(\theta,\phi) & = E_{q_{\phi}(z_x|x)}\Big\{ \left\| G_x(z_x,c) - x\right\|_1 + \left\| G_K(z_x,c) - K\right\|_1 \Big\} \nonumber \\ 
&+\lambda_{kl}  D_{\mathcal{KL}}( q_{\phi}(z_x|x) || \mathcal{N}(0,I)). 
\end{align}
\normalsize
By generating the keypoints as well as the images, it also encourages $z_x$ to more focusing on the semantic information of the keypoints. 
We can further encourage $z_x$ to be more pure towards semantics by 1) using relatively stronger decoder to generate images, and relatively weaker decoder to generate keypoints and 2) by weighting the importance or reconstructing keypoints and image.
The framework is illustrated as in Fig.~\ref{fig:framework} right.

The two variants of our model, multi-branch encoding and multi-branch decoding, both significantly improves the disentanglement and reconstruction quality in our experiments comparing with the baseline model of CVAE, especially for those challenging cases where the image-based CVAE fails (Fig.~\ref{fig:comp}). Our simple frameworks also outperform other more complicated methods by virtue of its more direct use of multi-view knowledge (Sec.~\ref{sec:exp}). The two frameworks have similar performances in most situations, and we will mainly show and discuss the results of the multi-branch encoding. It is important to note that the multi-view information such as keypoints and poses are directly extracted from the images; therefore no additional information is given. However, we show that by explicitly encouraging the latent code to model such information is a crucial step towards improving the model performance. 

\subsection{Detailed Implementation}
Our image-based CVAE model consists of an encoder and a decoder; each has six convolution or transposed convolution layers~\cite{dumoulin2016guide}. Both the convolution and the transposed convolution layers have the following parameters: kernel\_size = 4; stride = 2 and padding = 1. For the input, the images are randomly cropped and resized to 256x256. At the end of the encoder, the features are flattened as a 128-dim mean and variance vector, from which the 128-dim latent code $z_x$ is sampled using the re-parameterization trick. The domain label is represented using 1-hot, which is first mapped to another 128-dim code $z_c$ with a fully connected layer. During the decoding phase, $z_x$ is first concatenated with $z_c$ and is then decoded back to the image. The weights are initialized using Xavier~\cite{glorot2010understanding} except for the identity-conditioned fully connected layer, whose weights are sampled from $N\sim (0,1)$. By default, we use $\ell_1$ as reconstruction loss, and the weight of KL regularization is set to $\lambda_{kl}=0.1$.

For the multi-view model, the keypoints-based CVAE takes a 1-dim keypoints as input. The keypoint network also consists of an encoder and a decoder; each has four fully connected layers whose output channels are 500. The size of the latent code $z_{K}$ is also 128. We also use $\ell_1$ to measure keypoint reconstruction. For comparison between $z_K$ and $z_x$, we use $\ell_2$ to measure the distance. By default, the weight of the latent-consistency loss $\lambda_z$ is set to 1.

\section{Experiment Results}
\label{sec:exp}
We do experiments on two datasets: the Head-Mounted Display (HMD) dataset and the CMU Panoptic datasets~\cite{Joo_2017_TPAMI}; both of them have images and keypoints available. Our goal is to learn the identity-invariant latent representation, which can be applied to face/pose retargeting and reenactment. We compare against several methods regarding the quality of the output image and the learned latent code: the image-based CVAE baseline, UFDN~\cite{liu2018unified}, CycleGAN~\cite{zhu2017unpaired} and StarGAN~\cite{choi2017stargan}. Finally as an application, we also show how the latent code can be used to drive 3D avatars in VR and precisely transfer the facial expressions and eye movements. 

\subsection{Datasets}

\noindent\textbf{HMD Dataset} The Head-Mounted Display (HMD) dataset has images captured from cameras mounted on a head-mounted display (HMD). There are 123 different identities in total. For each identity, a diverse set of expressions and sentences are recorded in frames and are labeled as ``neutral face'', ``smile'', ``frown'', ``raise cheeks'' and others. For each frame, we capture different views such as the mouth, left eyes, and right eyes. During preprocessing, we sample the images near the peak frame of each expression, resulting in around 1,500 images for each identity. The images are gray-scale and are normalized and resized to 256x256 before given to the network as input. We also train a keypoint detector to extract the keypoints near the mouth, nose and the eyes.

\noindent\textbf{CMU Panoptic Dataset} We use the Range of Motion data in the CMU Panoptic dataset~\cite{ROM}. It consists of 32 identities, where each identity makes different poses under multiple VGA and HD cameras from different viewpoints. We use the images captured with the front-view HD camera, and each identity has around 7,000 images. The dataset also provides the 3D pose for each frame, which we transform and project to 2D to align with the images and use it as input.

\subsection{Cross-identity Image Translation}
\begin{figure}[t]
\centering
\small
\setlength\tabcolsep{1pt}
\begin{tabular}{r|ccccc}
 \includegraphics[width=.13\textwidth]{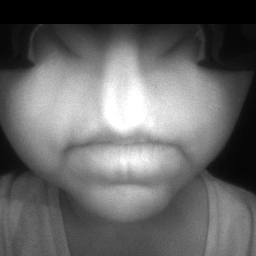} &
\includegraphics[width=.13\textwidth]{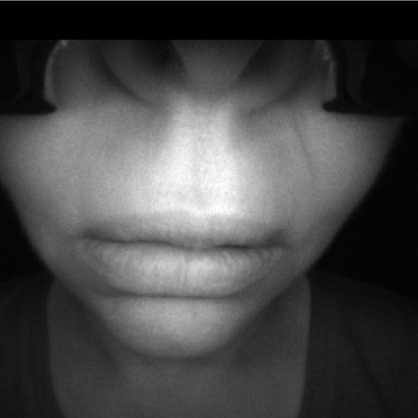}&
  \includegraphics[width=.13\textwidth]{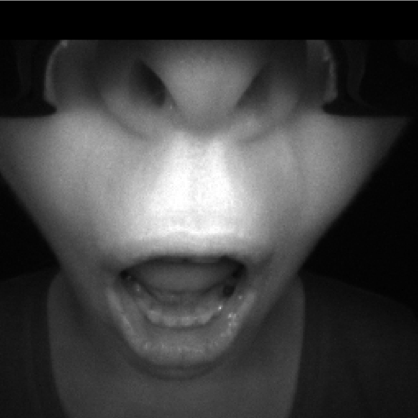}&
  \includegraphics[width=.13\textwidth]{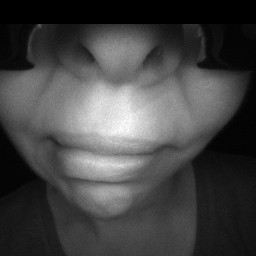}&
  \includegraphics[width=.13\textwidth]{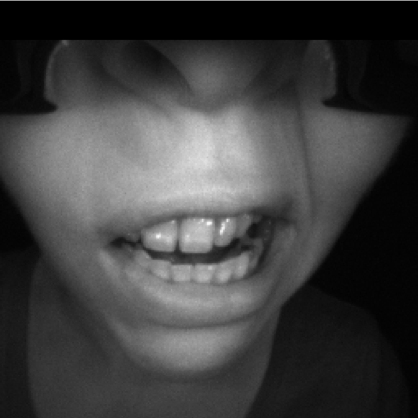}&
  \includegraphics[width=.13\textwidth]{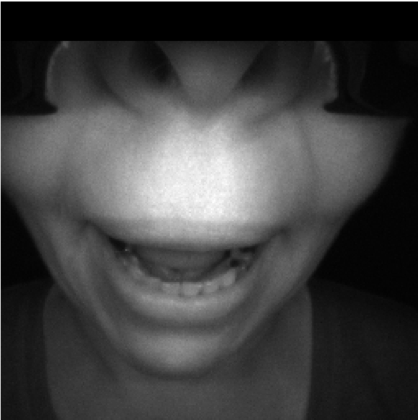}\\  \midrule
  {\tiny \textbf{Image CVAE}} &
  \includegraphics[width=.13\textwidth]{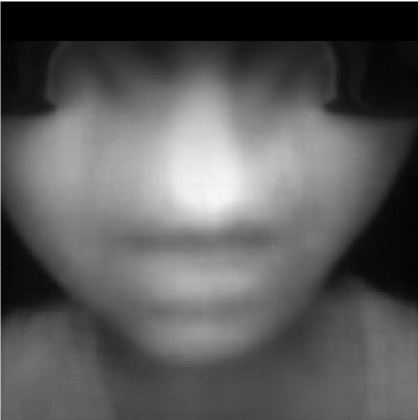}&
  \includegraphics[width=.13\textwidth]{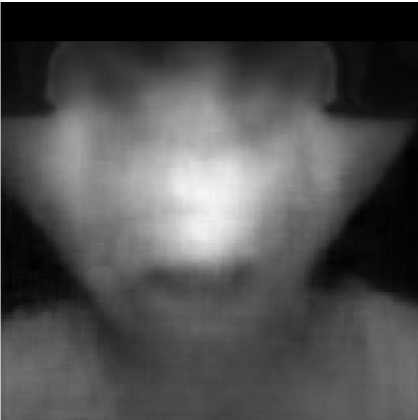}&
  \includegraphics[width=.13\textwidth]{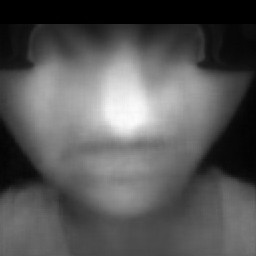}&
  \includegraphics[width=.13\textwidth]{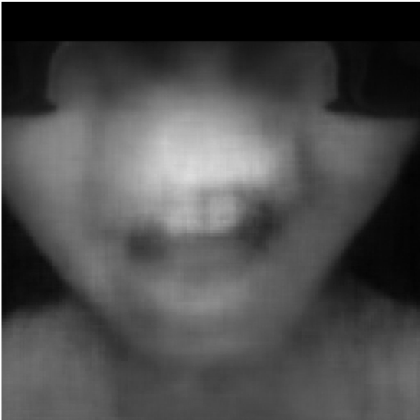}&
  \includegraphics[width=.13\textwidth]{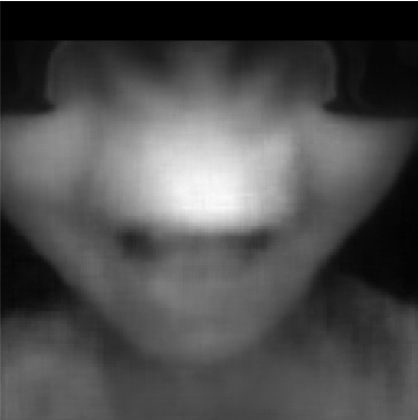}\\ 
  {\tiny \textbf{UFDN~\cite{liu2018unified}}}  &
  \includegraphics[width=.13\textwidth]{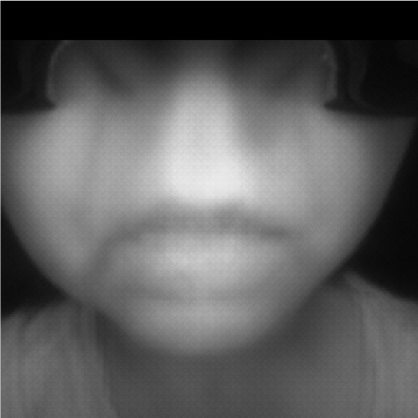}&
  \includegraphics[width=.13\textwidth]{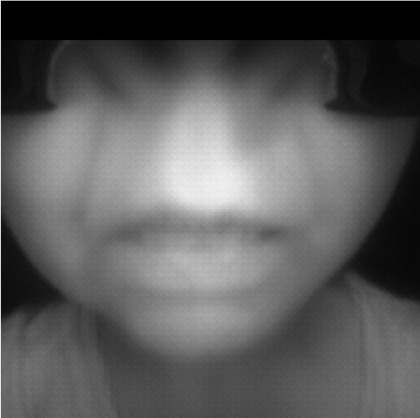}&
  \includegraphics[width=.13\textwidth]{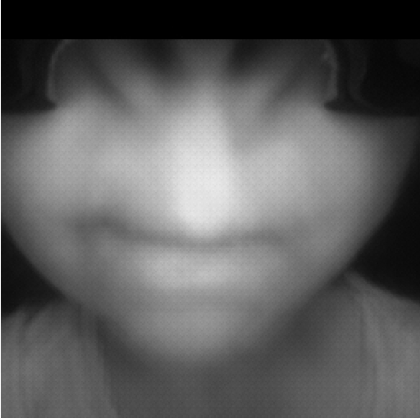}&
  \includegraphics[width=.13\textwidth]{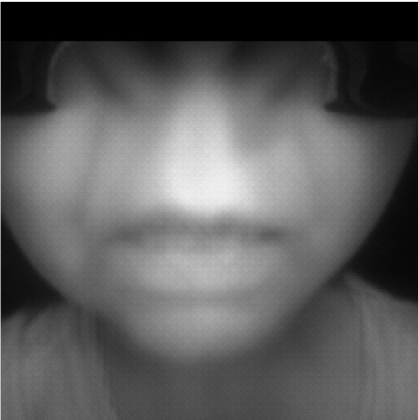}&
  \includegraphics[width=.13\textwidth]{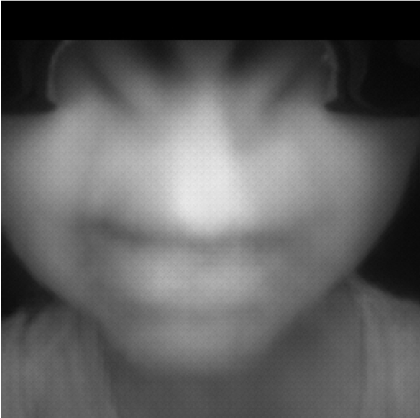}\\

   {\tiny \textbf{CycleGAN~\cite{zhu2017unpaired}}} &
  \includegraphics[width=.13\textwidth]{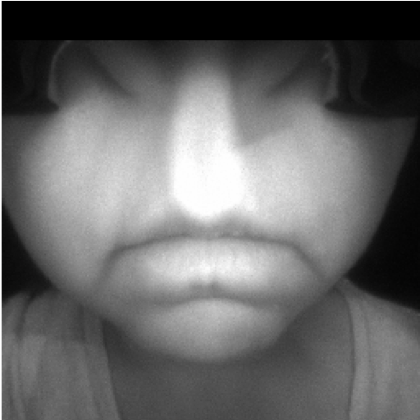}&
  \includegraphics[width=.13\textwidth]{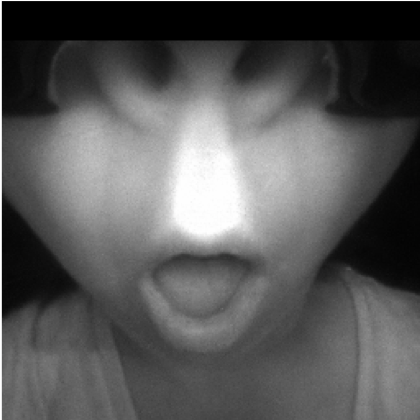}&
  \includegraphics[width=.13\textwidth]{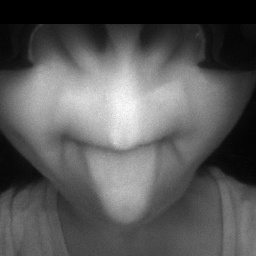}&
  \includegraphics[width=.13\textwidth]{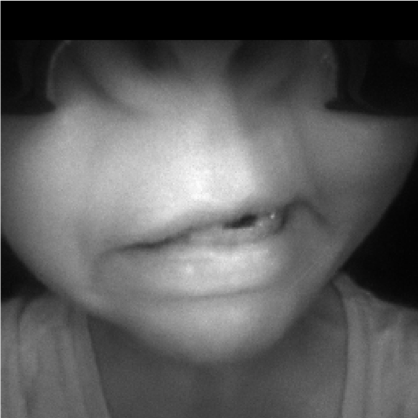}&
  \includegraphics[width=.13\textwidth]{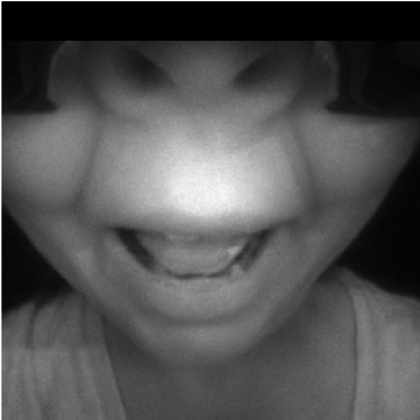}\\ 
{\tiny \textbf{StarGAN~\cite{choi2017stargan}}} &
  \includegraphics[width=.13\textwidth]{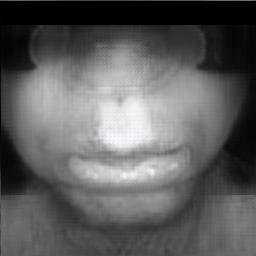}&
  \includegraphics[width=.13\textwidth]{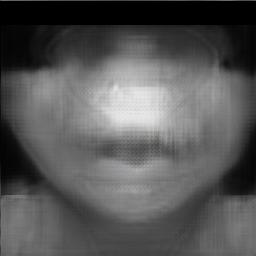}&
  \includegraphics[width=.13\textwidth]{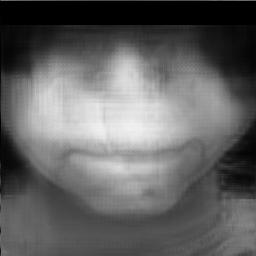}&
  \includegraphics[width=.13\textwidth]{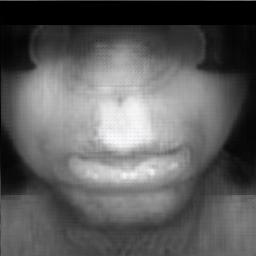}&
  \includegraphics[width=.13\textwidth]{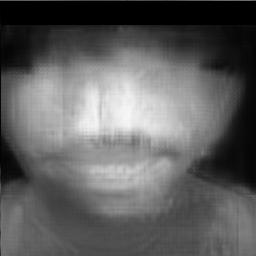} \\ 
{\tiny \textbf{Ours}} &
 \includegraphics[width=.13\textwidth]{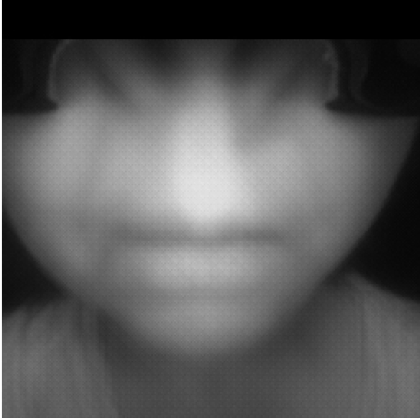}&
  \includegraphics[width=.13\textwidth]{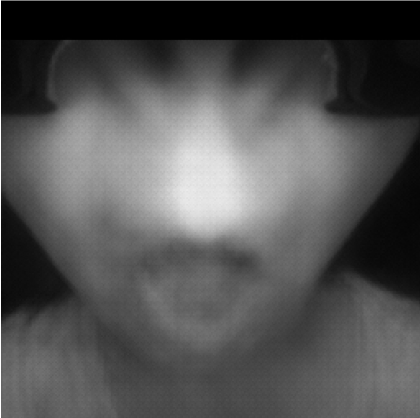}&
  \includegraphics[width=.13\textwidth]{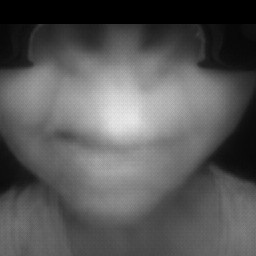}&
  \includegraphics[width=.13\textwidth]{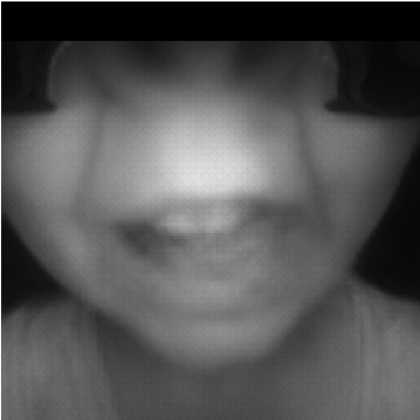}&
  \includegraphics[width=.13\textwidth]{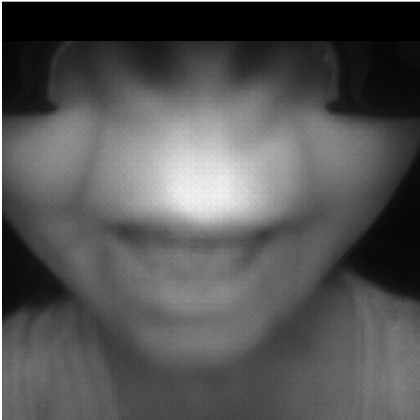}\\ 
\end{tabular}
\caption{Visual results of HMD image translation between two identities. The top row are the source id with different expressions (col 1-5) and the target id (col 6). Note how CycleGAN sometimes generates completely wrong expressions (e.g. row 4, col 3).}
\label{fig:hmd}
\end{figure}

\begin{figure}[t]
\centering
\small
\setlength\tabcolsep{1pt}
\begin{tabular}{r|ccccc}
\includegraphics[width=.13\textwidth]{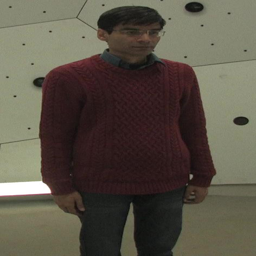} &
\includegraphics[width=.13\textwidth]{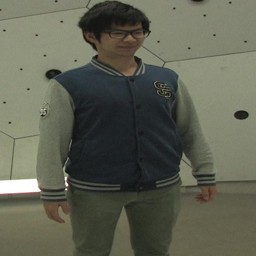}&
  \includegraphics[width=.13\textwidth]{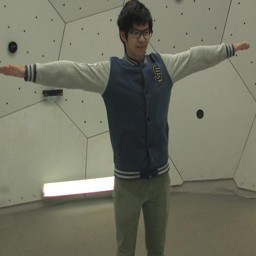}&
  \includegraphics[width=.13\textwidth]{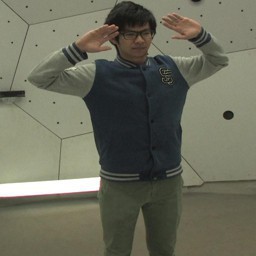}&
  \includegraphics[width=.13\textwidth]{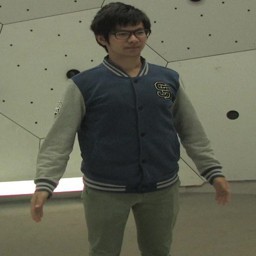}&
  \includegraphics[width=.13\textwidth]{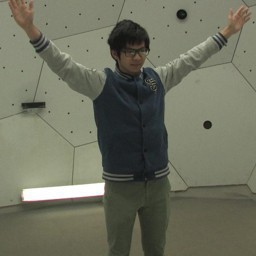}\\  \midrule
  {\tiny \textbf{Image CVAE}}  &
  \includegraphics[width=.13\textwidth]{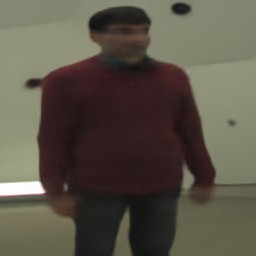}&
  \includegraphics[width=.13\textwidth]{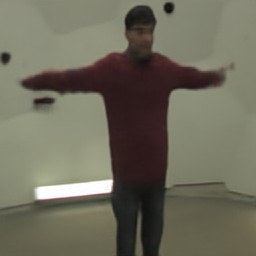}&
  \includegraphics[width=.13\textwidth]{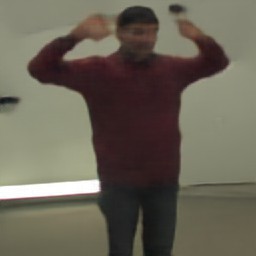}&
  \includegraphics[width=.13\textwidth]{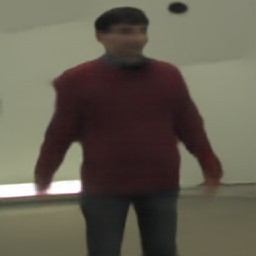}&
  \includegraphics[width=.13\textwidth]{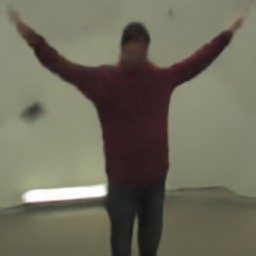}\\ 
  {\tiny \textbf{UFDN~\cite{liu2018unified}}}  &
  \includegraphics[width=.13\textwidth]{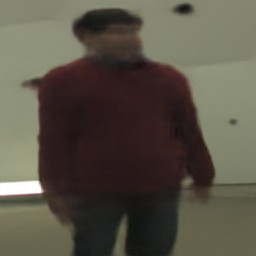}&
  \includegraphics[width=.13\textwidth]{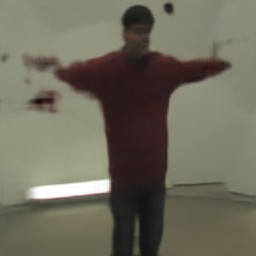}&
  \includegraphics[width=.13\textwidth]{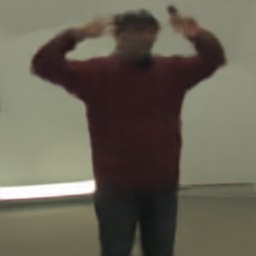}&
  \includegraphics[width=.13\textwidth]{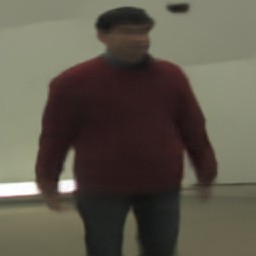}&
  \includegraphics[width=.13\textwidth]{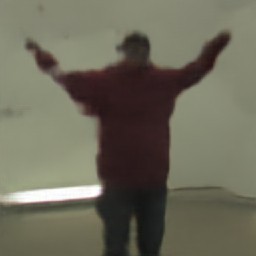}\\ 
  {\tiny \textbf{CycleGAN~\cite{zhu2017unpaired}}} &
  \includegraphics[width=.13\textwidth]{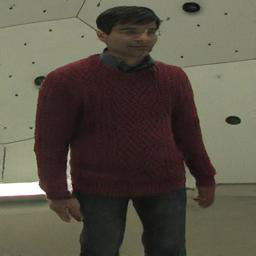}&
  \includegraphics[width=.13\textwidth]{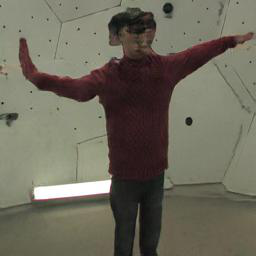}&
  \includegraphics[width=.13\textwidth]{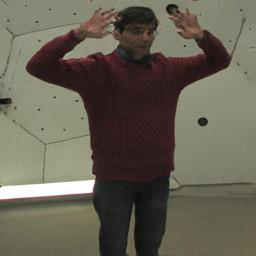}&
  \includegraphics[width=.13\textwidth]{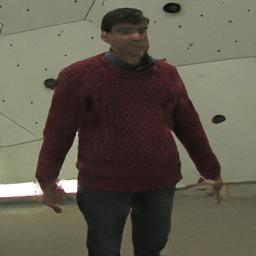}&
  \includegraphics[width=.13\textwidth]{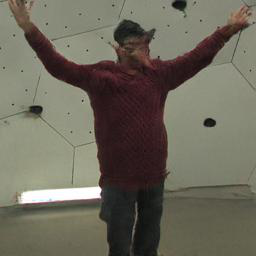}\\ 
  {\tiny \textbf{Ours}} &
 \includegraphics[width=.13\textwidth]{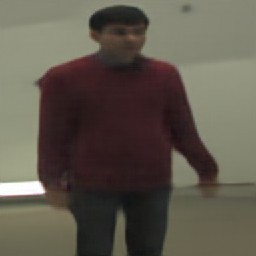}&
  \includegraphics[width=.13\textwidth]{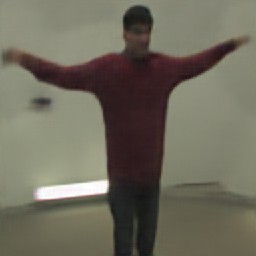}&
  \includegraphics[width=.13\textwidth]{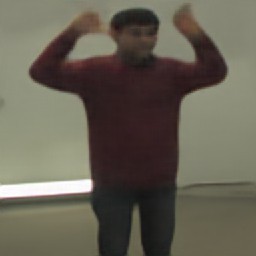}&
  \includegraphics[width=.13\textwidth]{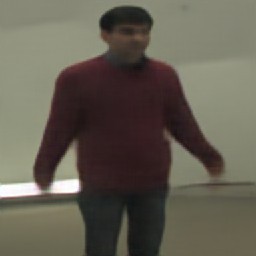}&
  \includegraphics[width=.13\textwidth]{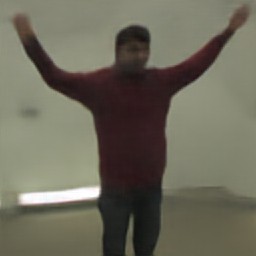}\\  
\end{tabular}
\caption{Visual results of Panoptic image translation. The top row are the source id with different poses (col 1-5) and the target id (col 6).}
\label{fig:pan}
\end{figure}

To achieve cross-identity image translation, we first train our multi-view CVAE model with all the identities. At test time, we only need to use the image-based CVAE component. Given an input image of source id, we first encode and sample the latent representation. Before decoding, we change the conditional id label to the target id to translate the image to the new identity while keeping the facial expressions. In Fig.~\ref{fig:hmd}, we show examples of HMD results between two identities and compares with other methods. As our baseline, the image-based CVAE does not deliver satisfying image quality nor preserves the semantics well. By applying adversarial training to the latent code~\cite{liu2018unified}, the quality improves while all expressions collapse to neutral faces. CycleGAN~\cite{zhu2017unpaired} generates sharp images given the adversarial training applied on the output image; although the semantics are not well preserved for some frames (column 3, 4 for example). CycleGAN is also limited to two domains and is not obvious how to infer the latent representations. StarGAN~\cite{choi2017stargan} as the multi-domain version of CycleGAN, produces fuzzier images with many artifacts. Our results, although not being as sharp as CycleGAN results, have fewer artifacts and more accurately preserves the expressions. Similar observations can be made from the results of Panoptic Datasets (Fig.~\ref{fig:pan}). Note that although we only showed results between two identities, CVAE, UFDN, StarGAN and ours are all multi-domain models and are trained and tested with all the identities in the dataset.

\begin{figure}[h]
\centering
\small
\setlength\tabcolsep{1pt}
\begin{tabular}{ccccc}
\includegraphics[width=.13\textwidth]{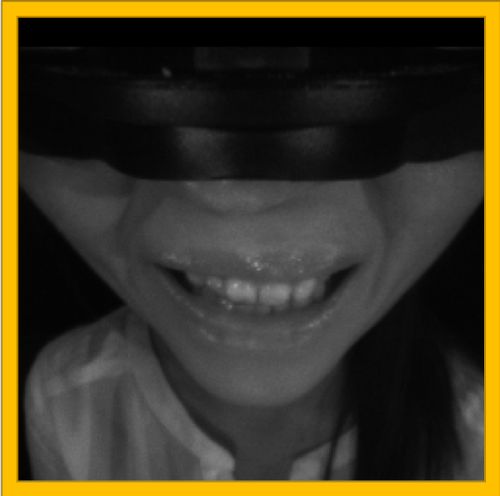}&
 \includegraphics[width=.13\textwidth]{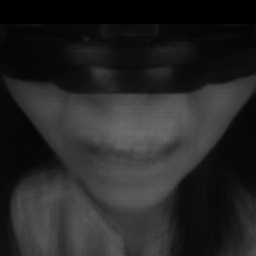}&
\includegraphics[width=.13\textwidth]{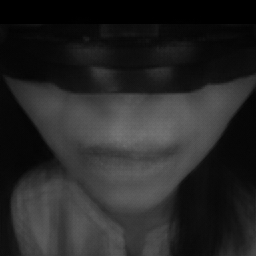}&
\includegraphics[width=.13\textwidth]{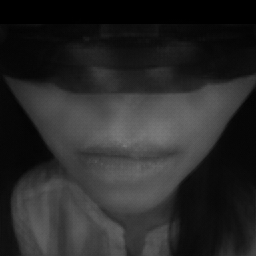}&
 \includegraphics[width=.13\textwidth]{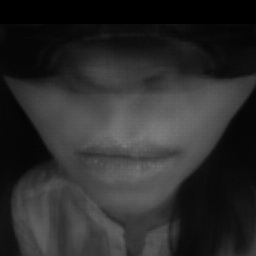}\\ 
 \includegraphics[width=.13\textwidth]{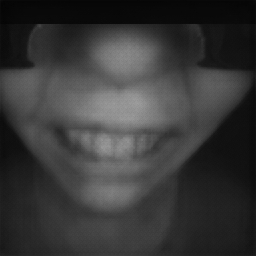}&
 \includegraphics[width=.13\textwidth]{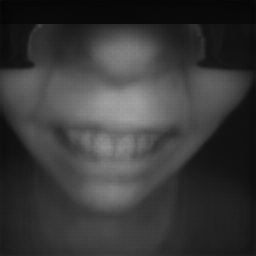}&
\includegraphics[width=.13\textwidth]{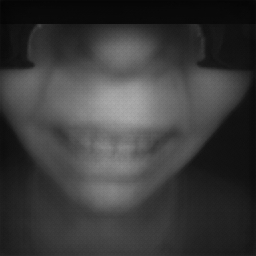}&
\includegraphics[width=.13\textwidth]{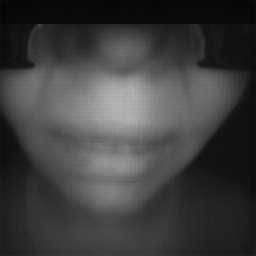}&
 \includegraphics[width=.13\textwidth]{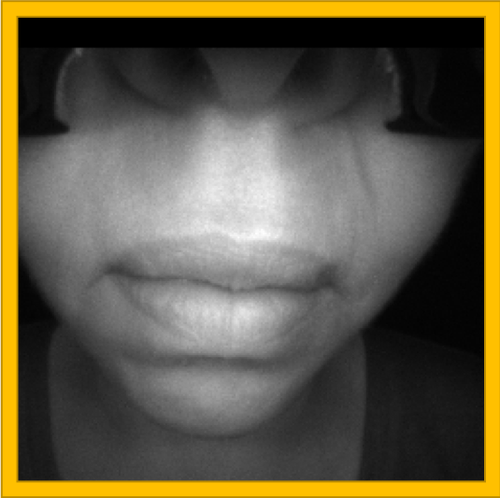}\\ 
\end{tabular}
\caption{Interpolation between expressions of two different ids. We mark the source images with colored borders. The rest are interpolation results.}
\label{fig:interp}
\end{figure}

Fig.~\ref{fig:interp} shows examples of expression interpolation between different identities. Given two source images of different ids, we extract the latent code of each image using the trained model. Since the codes are identity-invariant, we can interpolate between them as approximate representations of transitional expressions. We then concatenate the code with either identity label and decode to images to get identity-specific interpolated faces between the two source expressions.

\begin{figure}[h]
\centering
\small
\setlength\tabcolsep{1pt}
\begin{tabular}{ccccc}
\includegraphics[width=.13\textwidth]{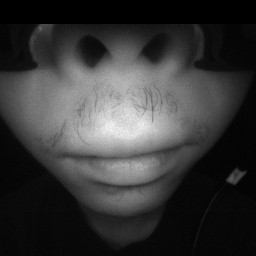}&
\includegraphics[width=.13\textwidth]{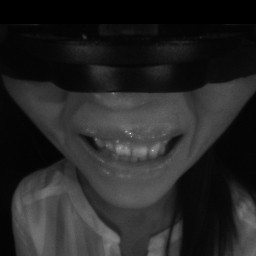}&
\includegraphics[width=.13\textwidth]{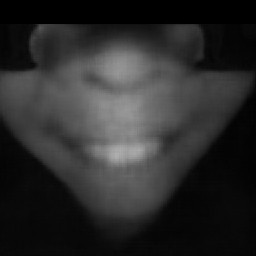}&
\includegraphics[width=.13\textwidth]{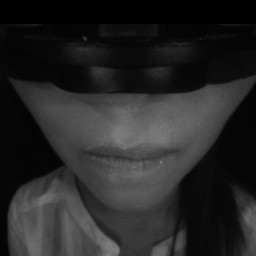}&
\includegraphics[width=.13\textwidth]{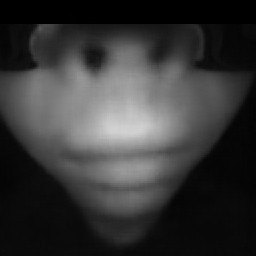}\\ 
(a) & (b) & (c) & (d) & (e) \\
\end{tabular}
\caption{Generalizing to new identities. (a) is the target person not seen during training. (b) and (d) are source expressions. (c) and (e) are the retargeted faces where the decoder takes the regressed id of (a) as the conditioning label.}
\label{fig:ni}
\end{figure}

For new identity not present in the training set, we cannot directly retarget since the model does not ``see'' the label corresponding to that identity during training.  This can be handled by taking the existing ids as the basis, and the new id can be viewed as a combination of existing ids. To achieve this, we learn a regressor from the training images to its 1-hot labels. We then regress the images of the new person to a combination of the existing 1-hot labels as the new label for decoding. Fig.~\ref{fig:ni} shows example results of face retargeting to a new identity.

We evaluate our results quantitatively using both unsupervised and supervised metrics: i) Auto-encoder (AE) error. We train separate VAEs for each identity. Given a translated image with the target identity $i$, we measure the reconstruction error after giving it as input to the $i_{th}$ VAE. The idea is that the trained VAEs should reconstruct the images better when the input is more similar to the images of the specific identity that it is trained on. ii) Classification error. We train an identity classification model and use it to compute the cross-entropy error of the translated image w.r.t. the target id.  i) and ii) both serve to evaluate the translated image quality and how similar it resembles the target domain, but do not measure how correctly it preserves the source semantics (expression and poses). For that purpose, iii) we use the available labelings of peak frames for the expressions and use those frames as the corresponding ground truth to directly measure the $\ell_2$ error. This metric measures both the image quality and the correctness of semantics. Results are shown in Table~\ref{table:numerical}.

Both CycleGAN and UFDN perform well on the first two metrics. This is expected as they tend to have high image quality, especially for CycleGAN. On the other hand, our results have the smallest $\ell_2$ error when comparing with the ground truth correspondence. This shows that our results not only have favorable quality but also best preserve the semantics. UFDN tends to lose semantics and generate mostly neutral faces, so their $\ell_2$ error is large. Comparing with the image-based CVAE baseline which does not use keypoints, our quality and semantics are both significantly better, showing the effectiveness of using multi-view training.

\begin{table}[h!]
\begin{center}
\resizebox{.5\textwidth}{!}{%
{
  \begin{tabular}{ l | c  c  c}
    \hline
    \textbf{Method} & \textbf{AE Error} &  \textbf{Classification Error} & \textbf{$\ell_2$ Error} \\ \hline
    \emph{Ground Truth} &         0.31 & 0    & 0 \\ 
    \emph{Image CVAE} & 1.62 & 2.60   & 1.15\\ 
    \emph{UFDN~\cite{liu2018unified}} &       0.56 & 1.10  & 2.21\\ 
    \emph{CycleGAN~\cite{zhu2017unpaired}} &   \textbf{0.33} & \textbf{0.18} & 0.82 \\ 
    \emph{StarGAN~\cite{choi2017stargan}} &    2.02 & 4.76  & 3.22 \\ \hline
    \emph{Ours} &       0.51 & 0.36 & \textbf{0.76} \\ \hline
    \hline
  \end{tabular}}
  }
  \end{center}
  \caption{Numerical comparisons. Our results have favorable quality (as shown in AE and classification error) and best preserve the semantics (as shown in $\ell_2$ error).}
  \label{table:numerical}
\end{table}

\begin{figure}[h!]
\centering
\small
\setlength\tabcolsep{1pt}
\begin{tabular}{cc}
\includegraphics[width=.44\textwidth]{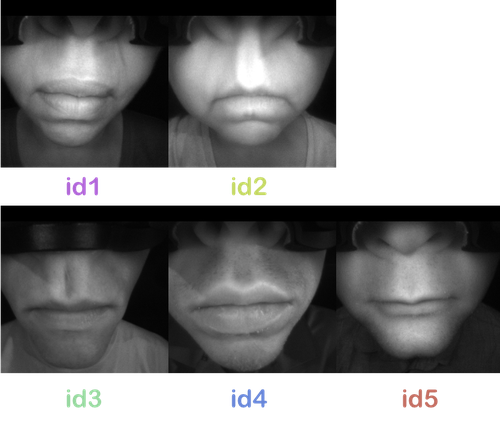}~~&
\includegraphics[width=.36\textwidth]{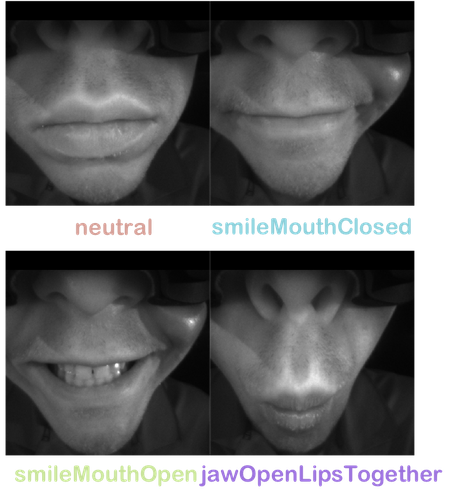}\\ %
\includegraphics[width=.44\textwidth]{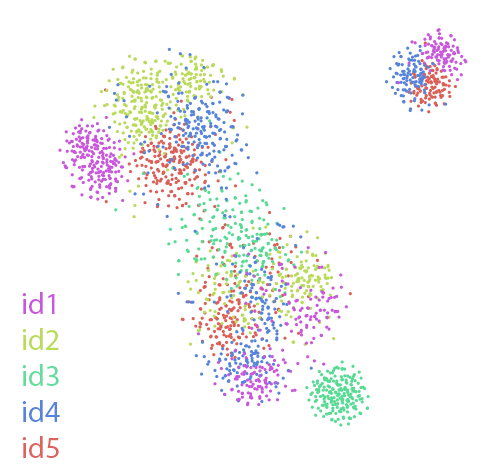}~~&
\includegraphics[width=.36\textwidth]{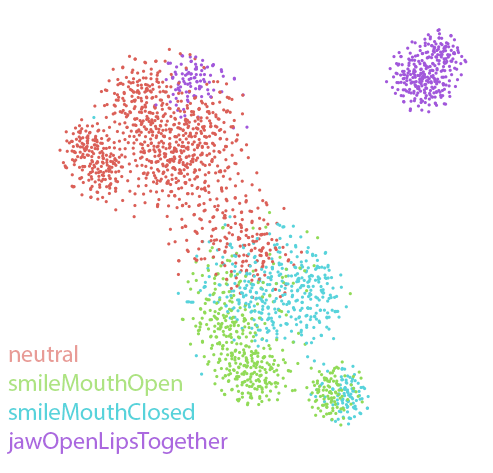}\\ 
\includegraphics[width=.36\textwidth]{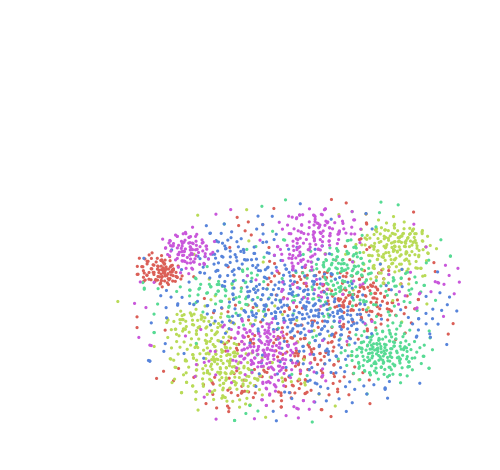}&
\includegraphics[width=.36\textwidth]{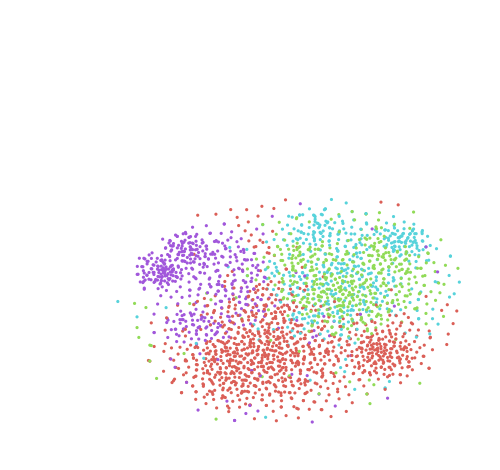}\\ 
\end{tabular}
\caption{t-SNE visualization of the embeddings of five identities and four expressions (shown above). t-SNE top: ours. t-SNE bottom: CVAE baseline. We color the points by both identities (left) and expressions (right).}
\label{fig:tsne}
\end{figure}

\subsection{Identity-Invariant Representations}
We visualize the latent representations using t-SNE to show their identity-invariance property (Fig.~\ref{fig:tsne}). We plot the latent codes of the images from five identities and four representative expressions, and color the points by both the expressions and the identities. We can see from the visualization that our representations (top) separate different expressions well, and also result in clusterings of the images in the same expression but from different identities. The distance between the points further shows that it is a meaningful metric as for how similar two expressions are from each other. For example, ``jaw open lips together'' is very different from other expressions, and their points are isolated from others. As for the image-based CVAE (bottom), the points of different expressions are largely mixed without clear separation.

\begin{figure}[t!]
\centering
\small
\setlength\tabcolsep{1pt}
\begin{tabular}{c}
\includegraphics[width=.6\textwidth]{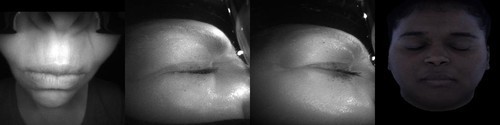}\\
\includegraphics[width=.6\textwidth]{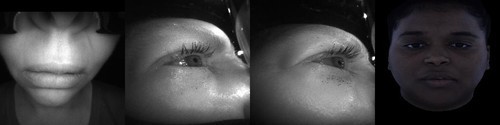}\\
\includegraphics[width=.6\textwidth]{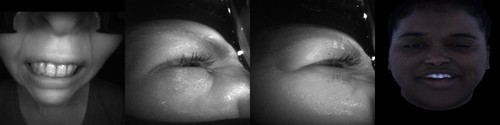}\\
\includegraphics[width=.6\textwidth]{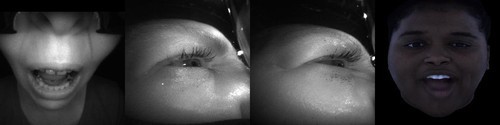}\\ 

\end{tabular}
\caption{Driving the 3D avatar of another person in VR while wearing the headset. From left to right: mouth, left eye, right eye, and the rendered 3D avatar of target identity (different from the source id wearing the headset). Note the left and right are mirrored between the image and the avatar.}
\label{fig:avatar}
\end{figure}

We show the application of our approach in social virtual reality (VR). Combining with ``eyes'', we train three multi-view CVAEs, one for the face, and two for the eyes. These three models encode the facial expressions and eye movements of any person to a shared latent space. Given the parameter of the 3D avatar of the target id at each frame and the corresponding image, we train a regression model to the 3D parameters of the avatar from the latent code encoded from the face and eye images of the target person. Then given another person wearing the headset, we can encode their face and eyes with the trained CVAE models and map them to the shared latent space, which is further translated to the avatar parameters with the trained regressor. Assuming the latent space are identity-invariant, it allows us to encode and transfer the facial expression and eye movement of any user to the target avatar. Fig.~\ref{fig:avatar} shows some examples of our results. 

\section{Conclusions}

We proposed the multi-view CVAE model to learn disentangled feature representations for data across multiple domains. Our model leverages multiple data sources, such as images, keypoints and poses, and formulate them as additional constraints when training the CVAE model. It explicitly guides the learned representation to encode the semantics that are shared across domains while leaving out the domain-specific attributes. We show our model can be applied to human retargeting and demonstrate the effectiveness of using additional ``views'' of data, which leads to improved reconstruction quality and better disentangling representations.

\section*{References}

\bibliography{mybibfile}

\end{document}